\documentclass[conference]{IEEEtran}
\IEEEoverridecommandlockouts
\usepackage{cite}
\usepackage{amsmath,amssymb,amsfonts}
\usepackage{algorithmic}
\usepackage{graphicx}
\usepackage{textcomp}
\usepackage{subcaption}
\usepackage{xcolor}
\usepackage[hidelinks]{hyperref}
\hypersetup{
    colorlinks=true,
    linkcolor=black,
    filecolor=green,
    urlcolor=green,
    citecolor=black,
}

\def\BibTeX{{\rm B\kern-.05em{\sc i\kern-.025em b}\kern-.08em
    T\kern-.1667em\lower.7ex\hbox{E}\kern-.125emX}}
\begin{document}

\title{Rethinking Early-Fusion Strategies for Improved Multimodal Image Segmentation\\
\thanks{This work was supported by the National Science and Technology Major Project under Grant 2020AAA0107300, Xinjiang Department and Office Linkage Project-Research on Key Technology of Self-Evolving and Learning Surface Mine Autopilot System-2023B01006, Xinjiang Key Laboratory of Intelligent Production and Control of Open Pit Mines (XJQY2007), the Postgraduate Research $\&$ Practice Innovation Program of Jiangsu Province (KYCX23$\_$2713), the Graduate Innovation Program of China University of Mining and Technology (2023WLKXJ100).}
}

\author{
    \IEEEauthorblockN{
    Zhengwen~Shen\qquad
    Yulian~Li \qquad 
    Han~Zhang \qquad 
    Yuchen~Weng\qquad
    Jun Wang$^{\star}$
    }
    School of Information and Control Engineering, \\China University of Mining and Technology, \\
Xuzhou, Jiangsu 221116, China.\\
    \thanks{
   *Corresponding author: Jun Wang (jrobot@126.com).
    }
}

\maketitle

\begin{abstract}
RGB and thermal image fusion have great potential to exhibit improved semantic segmentation in low-illumination conditions. Existing methods typically employ a two-branch encoder framework for multimodal feature extraction and design complicated feature fusion strategies to achieve feature extraction and fusion for multimodal semantic segmentation. However, these methods require massive parameter updates and computational effort during the feature extraction and fusion. To address this issue, we propose a novel multimodal fusion network (EFNet) based on an early fusion strategy and a simple but effective feature clustering for training efficient RGB-T semantic segmentation. In addition, we also propose a lightweight and efficient multi-scale feature aggregation decoder based on Euclidean distance. We validate the effectiveness of our method on different datasets and outperform previous state-of-the-art methods with lower parameters and computation.
\end{abstract}


\section{Introduction}
\label{sec:intro}
Semantic segmentation is critical in computer vision tasks that enable pixel-by-pixel dense scene perception and understanding. Most existing methods are focused on the semantic segmentation of RGB images under normal illumination conditions and achieved excellent results in segmentation accuracy\cite{ xie2021segformer}. However, these methods significantly decrease segmentation performance when dealing with RGB images under adverse illumination conditions. Therefore, to address the adverse effects of illumination changes, the research on RGB-T semantic segmentation utilizing thermal images to provide stable supplementary information for RGB images has received increasing attention\cite{ha2017mfnet, shivakumar2020pst900, shen2024ecfnet, shen2024hefanet}.

The main challenge of multimodal semantic segmentation is to effectively fuse multimodal features to utilize the advantages of each modality fully. Earlier methods typically achieved fusion through simple pixel summation or channel concatenation, which often resulted in coarse feature information and introduced noise\cite{ha2017mfnet, sun2019rtfnet}. 
\begin{figure}[h]
\centering
\includegraphics[width=\linewidth]{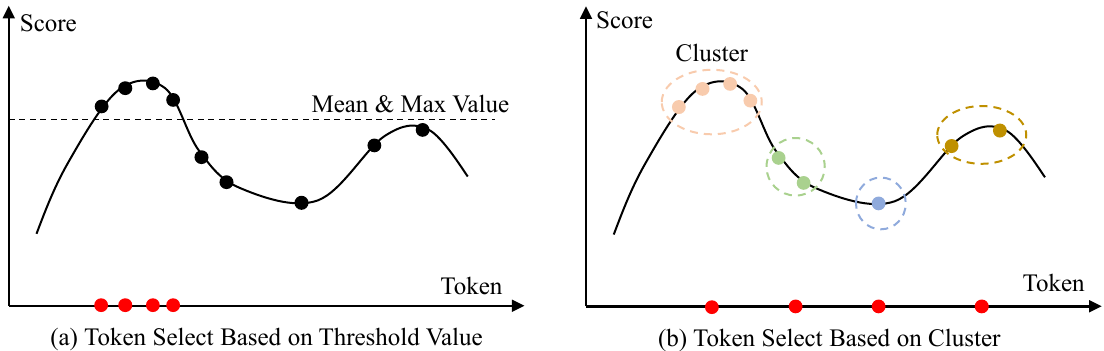}
\caption{Feature selection based on thresholds or clustering.}
\label{fig:issue}
\end{figure}
Recently, most methods have employed more sophisticated fusion strategies, often introducing attention mechanisms to reweight important multimodal features\cite{wang2023sgfnet, li2022rgb, zhou2021gmnet,zhang2023cmx}. However, with the increase of the depth of the feature extraction network, the number of parameters for feature extraction increases significantly, and the complex fusion strategy also considerably increases the computational burden and does not always significantly improve performance. These problems may be caused by the feature difference between RGB and thermal images, which leads to the feature extraction network being biased towards a certain modality during continuous training. Considering these issues, a natural question arises: \textit{is there a superior method that can maintain high training efficiency, low parameter, and computational load, and still perform effective multimodal semantic segmentation?}

To answer the above questions, we rethink the multimodal early fusion strategy to construct an efficient multimodal feature fusion network. Specifically, we first propose a multimodal feature interaction and fusion module to alleviate the differences between modalities. Considering that early fusion has integrated multimodal information, the next key question is how to effectively process the fusion information in the feature extraction process to avoid the features of a certain modality being dropped in the downsampling process of the cascading network. As shown in Fig~\ref{fig:issue}(a), the pooling operation is usually used for the downsampling process of the feature map, but this approach samples the mean or maximum value, which tends to overlook the information that may be important in the subsequent stages. For this reason, as shown in Fig~\ref{fig:issue}(b), motivated by \cite{zeng2022not}, we propose a token clustering sampling method based on a balanced strategy of semantic and spatial distances, which ensures that the sampling stage retains rich semantic feature information. In this way, our proposed EFNet can effectively improve the RGB-T semantic segmentation performance with lower parameters and computational effort and can be applied to multiple available RGB-T semantic segmentation datasets.

The main contributions of this paper are as follows:
\begin{itemize}
\item We propose a novel multimodal fusion network based on an early fusion strategy efficiently applied to RGB-T image semantic segmentation.
\item We propose a multimodal feature interaction fusion module based on feature separation and reconstruction to achieve early interaction and fusion of two modal feature information. In addition, we propose a semantic distance and spatial distance clustering module based on a balanced strategy to implement feature downsampling efficiently.
\item We propose a lightweight decoder based on Euclidean distance for feature aggregation to achieve aggregation and reconstruction of multiscale fused features.
\end{itemize}

\begin{figure}[!h]
\centering
\includegraphics[width=\linewidth]{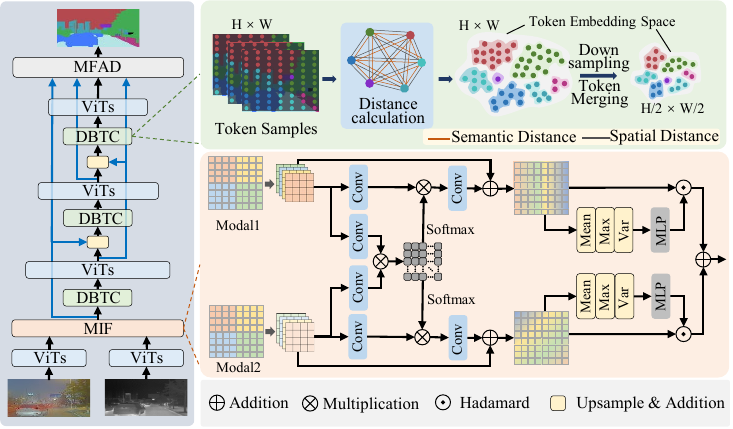}
\caption{Overall architecture of the proposed method. }
\label{fig:framework}
\end{figure}

\section{PROPOSED METHOD}
\label{sec:PROPOSED METHOD}

\subsection{Overview Architecture}
As shown in Fig~\ref{fig:framework}, given the RGB and thermal images, the feature maps of the first stage are extracted by transformer blocks, and the obtained multi-modal features are input to the Multimodal Feature Interaction and Fusion (MIF) module to obtain interactive fusion features. Then, the fusion feature map is input to the Dual-distance Balanced Token Clustering (DBTC) module for cluster sampling, and the obtained downsampled image is input to the next ViTs block. In addition, before the third and fourth stages of feature extraction, we downsample the features of the previous layer, sum the elements with the feature map of the current layer, and then input them into the DBTC module. Finally, the multi-scale features obtained at each stage are input into the Multi-scale Feature Aggregation Decoder (MFAD) to predict the results.

\subsection{Multimodal Feature Interaction and Fusion}
Given the high resolution of the input features, direct attention computation imposes a substantial computational burden. To address this, motivated by\cite{tu2022maxvit}, the MIF module divides the high-resolution feature map of each modality into smaller windows and performs cross-attention modeling. This approach significantly reduces the number of pixels that need to be processed for each attention operation, thereby lowering the computational cost. Additionally, after attention modeling, these windows are restored to the original feature resolution to maintain the integrity of the feature information.

Specifically, we divide the feature maps of $F_{R}$ and $F_{T}$ into n windows $W_{R}^{n}$ and $W_{T}^{n}$, respectively. Then, we multiply the obtained multimodal features by the matrix to obtain the multimodal interaction feature matrix $I_{R}^{n}$ and $I_{T}^{n}$. Although adopting window partitioning can significantly reduce the computational burden of high-resolution features, it lacks the interaction between different windows. To solve this problem, we introduce a channel attention mechanism based on statistical computation, which maps global statistical results, such as average pooling, maximum pooling, and variance, through multiple perception layers to obtain channel attention scores. Based on these obtained scores, we reweight the interaction features in different window partitions and add elements to the resulting modal features.

\begin{table*}[!t]
\caption{RGB-T semantic segmentation results on MFNet dataset.}\small
\centering
\resizebox{\linewidth}{!}{\small\begin{tabular}{c c c c c c c c c c c c}\\
\hline
&RTFNet &ABMDRNet &MTANet &RSFNet &SFAF-MA &GMNet &LASNet &SGFNet &MDRNet & Ours\\ 
&\cite{sun2019rtfnet} &\cite{zhang2021abmdrnet}  &\cite{zhou2022mtanet} &\cite{li2024residual} &\cite{he2023sfaf}&\cite{zhou2021gmnet} &\cite{li2022rgb} &\cite{wang2023sgfnet} &\cite{zhao2023mitigating} &\\
\hline\hline
\multicolumn{1}{c}{Backbone} &ResNet-152 &ResNet-50 &ResNet-152 &ResNet-50  &ResNet-50 &ResNet-152 &ResNet-152 &ResNet-101 &ResNet-50 & Tcformer-L
\\
\multicolumn{1}{c}{Params (M)} &337.1  &64.6  &121.9 &73.63 & 156 &149.8 &139.9 &125.3 &64.6 & \textbf{29.5}
\\
\multicolumn{1}{c}{FLOPs (G)} & 337.0 & 194.3 & - & 98.8 & 354 & - & 233.8 & 143.7 & 194.3 & \textbf{36.6}
\\
\multicolumn{1}{c}{mIoU (\%)} &53.2 &54.8 &56.1 &56.2 & 55.5 &57.3 &54.9 &57.6 &56.8 & \textbf{57.9}
\\
\hline
\end{tabular}}
\label{tb:MFNet}
\end{table*}

\begin{table*}[!t]
\caption{RGB-T semantic segmentation results on PST900 dataset.}\small
\centering
\resizebox{\linewidth}{!}{\small\begin{tabular}{c c c c c c c c c c c}\\
\hline
&RTFNet &PSTNet &MTANet & GMNet &EGFNet &LASNet &MDRNet &MMSMCNet &SGFNet & Ours\\ 
&\cite{sun2019rtfnet} &\cite{shivakumar2020pst900} &\cite{zhou2022mtanet} &\cite{zhou2021gmnet} &\cite{zhou2022edge} &\cite{li2022rgb} &\cite{zhao2023mitigating} &\cite{zhou2023mmsmcnet}&\cite{wang2023sgfnet} &\\
\hline\hline
\multicolumn{1}{c}{Backbone} &ResNet-152 &ResNet-18 &ResNet-152 &ResNet-50 &ResNet-152 &ResNet-152  &ResNet-50 &MiT-B3 &ResNet-101 & Tcformer-L
\\
\multicolumn{1}{c}{Params (M)} &337.1  &105.8 &121.9 & 149.8 & 201.3  & 93.5 & 64.6 & 98.6 &125.6 & \textbf{29.5}
\\
\multicolumn{1}{c}{mIoU (\%)} &60.6 &68.4 &78.6 & 84.1 &78.5 &84.4 &74.6 & 79.8 &82.8 & \textbf{85.6}
\\
\hline
\end{tabular}}
\label{tb:PST900}
\end{table*}

\begin{table*}[!t]
\caption{RGB-T semantic segmentation results on FMB dataset.}\small
\centering
\small\begin{tabular}{c c c c c c c c}\\
\hline
&RTFNet &EGFNet &LASNet &GMNet &SFAF-Net &EAEFNet & Ours\\ 
&\cite{sun2019rtfnet} &\cite{zhou2022edge} &\cite{li2022rgb} &\cite{zhou2021gmnet} & \cite{liang2023explicit} &\cite{he2023sfaf} &\\
\hline\hline
\multicolumn{1}{c}{Backbone} &ResNet-152 &ResNet-152 &ResNet-152 &ResNet-50  &ResNet-152 &ResNet-50& Tcformer-L
\\
\multicolumn{1}{c}{Params (M)} &337.1 &201.3  &139.9 &149.8 &156 & 77.93&\textbf{29.5}
\\
\multicolumn{1}{c}{mIoU (\%)} &42.7 &47.3  &53.4 &49.2 &42.7 & 53.5 &\textbf{55.2}
\\
\hline
\end{tabular}
\label{tb:FMB}
\end{table*}
\subsection{Dual-distance Balanced Token Clustering}
From the perspective of the distribution space of image feature information, adjacent pixels tend to show higher similarity. Based on the above observations, we propose a new token clustering method based on a double-distance balancing strategy Specifically, we calculate the semantic distance and spatial location distance between tokens, denoted as $d_{i,j}$, and defined as follows:
\begin{equation}\label{eq6}
\begin{split}
d_{i,j}=\left \| x_{i}-x_{j} \right \|_{2}+ \left( 1-\tau  \right) \left \| y_{i}-y_{j} \right \|_{2}
\end{split}
\end{equation}
where $x_{i}$ and $x_{j}$ represent the $i-th$ and $j-th$ token features of semantic feature set $x$, $y_{i}$ and $y_{j}$ are the corresponding spatial location set $y$, and $\tau$ is the weighted factor for semantic and spatial distance.
Then, we compute the local density $\rho$ for each token based on its k-nearest neighbors, as shown in equation~\ref{eq2}:
\begin{equation}\label{eq2}
\begin{split}
\rho_i=\exp{\left(-\frac{1}{k}\sum_{x_j}d_{i,j}^{2}\right)}, {x_j\in KNN\left(x_i\right)}
\end{split}
\end{equation}
where \textit{KNN}$\left(x_i\right)$ denotes the k-nearest neighbors of \textit{i-th} token. For \textit{i-th} token, we adopt the distance metric $\delta_i$ as the minimum distance between it and any other tokens with a higher local density. In addition, for a token with the highest local density, its metric is set to the maximum distance between that token and all other tokens, the calculation is as follows:
\begin{equation}\label{eq8}
\delta_{i}=
\left\{
\begin{aligned}
    & \min \left( d_{i,j}\right), \exists j :\rho_{j}>\rho_{i} \\
    & \max \left( d_{i,j}\right), other
\end{aligned}
\right.
\end{equation}
The score for each token is calculated as $\rho_{i}\times \delta_{i}$, which reflects the likelihood of the token becoming a cluster center. Once the cluster centers are selected, other Tokens are assigned to the nearest cluster center based on their distance from the existing cluster centers. For token merging, we introduce a weighted fraction \textit{p} obtained by linear mapping to represent the weight of each token, and the merged tokens are as follows:
\begin{equation}\label{eq9}
X_i^*=\frac{\sum x_j e^{p_j} }{\sum e^{p_j} },j\in C_i,p_{j}\in P
\end{equation}
\begin{equation}
\label{eq10}
F=Softmax \left( \frac{Q\times K^T}{\sqrt{d_c}}+P\right) V
\end{equation}
where $C_i$ is the set of tokens in $i-th$ cluster, $p_j$ is the score of $x_j$. Subsequently, the original feature map $x_{i}$ and the merged feature map $X_{i}^*$ are added to feed into the next-stage transformer module. We added the importance score \textit{P} to the attention score to ensure that the important markers contribute more to the final output $F$, and $d_{c}$ is the number of channels in the current query vector.

\subsection{Multi-scale Feature Aggregation Decoder}
To better aggregate the multi-scale feature map, we propose a novel token aggregation decoder based on Euclidean distance, which is specially used to utilize the semantic distance information in features. It constructs learnable semantic class tags that act as anchors for each class in the feature space, effectively utilizing the multi-scale information and semantic distance presented in high-resolution feature maps. By calculating the distance between each tag feature and the semantic class tag, the decoder accurately assigns each pixel to its corresponding class, enhancing the overall performance of semantic segmentation. The process is defined as follows:
\begin{equation}
\label{eq11}
    X_{f}=Concat(Up(S^n)), n=1,2,3,4 \\
\end{equation}
\begin{equation}
\label{eq12}
    d_{i,j}^s=\left \| x_i^{sc}-x_j \right\|_2 , d_{i,j}^{s}  \in D^{s} \\
\end{equation}
\begin{equation}
\label{eq13}
    S=Softmax(-D^s)
\end{equation}
where $S^n$ denotes the feature map in the $n-th$ Transformer block. "Up" indicates that the clustering features of each block are upsampled to $H/4\times W/4$ resolution. $x_j$ denotes the $j-th$ token of $x$, $x_i^{sc}$ represents the learnable semantic category Token, and $d_{i,j}^{s}$ represents the semantic distance between the $j-th$ pixel feature $x_j$ and the class $i$ semantic category feature $x_i^{sc}$, and S is the final prediction. Since the Softmax function is an increasing function, a smaller median value of $D^{s}$ means that the closer the distance is, the smaller the resulting classification probability, which is the opposite of the goal, so we assign $D^{s}$ to a negative number.
\section{EXPERIMENTAL RESULTS}
\label{sec:EXPERIMENTAL RESULTS}

\subsection{Datasets}
We conducted comprehensive experiments on three public datasets:

\textbf{MFNet}\cite{ha2017mfnet}: consisting of 1569 pair images, divided by the dataset into 784 and 393 for training and testing.

\textbf{PST900}\cite{shivakumar2020pst900}: consisting of 894 pair images, which divided the dataset into 597 and 288 for training and testing.

\textbf{FMB}\cite{liu2023segmif}: consisting of 1500 pair images, which divided the dataset into 1220 and 280 for training and testing. 
\subsection{Implementation details}
The EFNet method is built upon pre-trained TCFormer-Light~\cite{zeng2022not}. The experiments implemented are based on PyTorch 1.8, which is trained on a Nvidia Tesla V100 GPU. We utilize the AdamW optimizer with a weight decay of 0.01 and a base learning rate of $6\times10^{-5}$. We used cross-entropy loss for training and mean Intersection-overUnion (mIoU) as the evaluation metric. The resolution size of the training phase is $640\times480$, and the batch size is 2.

\subsection{Comparisons with the state-of-the-arts}
We conducted experiments on three RGB-T datasets: MFNet, FMB, and PST900. Baseline (addition instead of MIF, no positional encoding, MLP as decoder) results: mIoU is $54.5\%$, $51.6\%$, and $80.4\%$, respectively.
Table~\ref{tb:MFNet} shows the comparison results of our proposed method with other advanced RGB-T semantic segmentation methods on the MFNet dataset, and we can conclude that our method performs well. Our method mIoU reached $57.9\%$, 0.3 percentage points higher than SGFNet. However, compared with SGFNet, the proposed method significantly reduces the number of parameters by 95.8M (\textbf{29.5M \emph{v.s.} 125.3M}) and the computational effort by 107.1G (\textbf{36.6G \emph{v.s.} 143.7G}), which fully proves the effectiveness of our proposed method.
As shown in Table ~\ref{tb:PST900} on the PST900 dataset, our method achieves the optimal result, with a mIoU result of $85.6\%$. In addition, as shown in Table ~\ref{tb:FMB}, the proposed method achieves $55.2\%$ on the mIoU value on the recently published FMB dataset and further verifies the effectiveness of the proposed method with fewer parameters.

In summary, the experimental results on multiple datasets fully demonstrate that the proposed method has excellent performance and high generalization, and shows great potential and practicability in RGB-T semantic segmentation tasks.
\begin{table}[h]%{0.3\textwidth}
        \centering
        \caption{\centering Ablation studies for fusion methods.}\small
        \renewcommand\tabcolsep{10pt}
        \renewcommand{\arraystretch}{1.2}
        \small\begin{tabular}{c c c} 
        \hline
        {Position} &{Params(M)} &{mIoU (\%)}
        \\
        \hline\hline
        Cat  &{29.56}  &54.0 \\
        Add  &{29.46}  &56.7 \\
        MIF  &{29.54}  &{57.9} \\
        \hline
       \end{tabular}
       \label{tb:Ablation(a)}
       %}
\end{table}
\begin{table}[h]%{0.33\textwidth}
        \centering
        \caption{\centering Ablation studies for the position encoding.}\small
        \renewcommand\tabcolsep{10pt}
        \renewcommand{\arraystretch}{1.2}
        \small\begin{tabular}{c c c} \hline
        Position &{Params(M)} &{mIoU (\%)}
        \\
        \hline\hline
        {w/o}  &{29.54}  &55.9 \\
        {PE}  &{29.55}  &57.0\\
        {PCE}  &{29.54}  &57.9 \\
        \hline
        \end{tabular}
        \label{tb:Ablation(b)}
\end{table}

\subsection{Ablation Study}
In this section, we performed a series of ablation studies on the MFNet dataset to explore how different architecture components affect segmentation performance.

\indent \textbf{Ablation study on the MIF:} We compare the proposed MIF module with other fusion operations, as shown in Table~\ref{tb:Ablation(a)}. ``Cat'' represents the concatenation operation, and ``Add'' represents an additive operation. Experimental results show that the proposed MIF module can better increase the interaction and fusion of modalities, slow down the differences between modalities, and improve the ability of feature joint representation. In addition, compared with other methods, the number of parameters in MIF is only increased by 0.08M, and the segmentation score reaches $57.9\%$.
\begin{table}[htp]
\centering
\caption{Ablation study of weight factor $\tau$ in DBTC.}
\renewcommand\tabcolsep{10pt}
\renewcommand{\arraystretch}{1.2}
\small\begin{tabular}{ccc}
\hline 
Weight $\tau$ & mAcc(\%)$\uparrow$ & mIoU(\%)$\uparrow$ \\
\hline 
1.0,  0.7,  0.3 & 65.7 & 56.1   \\
0.5,  0.5,  0.5 & 71.2 & 56.9   \\
0.3,  0.7,  1.0 & 66.9 & \textbf{57.7}   \\
\hline 
\label{tab:factor}
\end{tabular}
\end{table}
\begin{table}[h]%{0.33\textwidth}
        \centering
        \caption{\centering Ablation studies of the decoder.}\small
        \renewcommand\tabcolsep{10pt}
        \renewcommand{\arraystretch}{1.2}
        \small\begin{tabular}{c c c} \hline
        {Method} &{Params(M)} &{mIoU (\%)}
        \\
        \hline\hline
        Deeplab\cite{chen2017deeplab} &33.28  &57.1
        \\
        {MLP\cite{xie2021segformer}} &30.06  &56.3
        \\
        {Ours} &29.54 &57.9
        \\
        \hline
        \end{tabular}
        \label{tb:Ablation(c)}
\end{table}

\indent \textbf{Ablation studies on spatial position:}. We propose a new learnable pixel coordinate encoding (PCE) designed to calculate the distance between tokens during clustering. To this end, we performed an ablation experiment to verify its effectiveness, and the results are shown in Table~\ref{tb:Ablation(b)}. Compared with the method without PCE, the proposed method can improve the mIoU indicator to $57.9\%$, and the efficiency of pixel coordinate encoding is better than that of the common position coding (PE).

\indent \textbf{Ablation studies on DBTC:}. We further validated the weight factors $\tau$ configuration of the token clustering method. The experimental results are recorded in Table~\ref{tab:factor}, which shows that the model achieves optimal performance when the weight factors $\tau$ in the three DBTC modules are sequentially set to 0.3, 0.7, and 1.0. This result aligns with expectations: in the shallow network layer, features primarily capture local spatial information, crucial for clustering. In the deeper layer, as features encode rich semantics, clustering naturally shifts to rely more on semantic distance, with spatial distance playing a diminished role.

\indent \textbf{Ablation studies on different decoder}. We performed ablation studies on different decoder settings to verify the effectiveness of our proposed decoder, and the results are shown in Table~\ref{tb:Ablation(c)}. Compared to Deeplab and MLP decoders, our
proposed decoder achieves a higher segmentation score, with a mIoU of $57.9\%$ with few parameters introduced. This significant performance improvement confirms the advantages of our proposed semantic distance-based decoder in utilizing semantic distance information from token features.

\section{CONCLUSION}
\label{sec:CONCLUSION}
In this paper, we propose a novel simple, and efficient RGB-T semantic segmentation network. To overcome the high computational cost and large number of parameters required by existing methods, we adopt an early fusion strategy combined with a simple and effective feature clustering sampling method to achieve efficient RGB-T semantic segmentation. In addition, we introduce a lightweight multi-scale feature aggregation decoder based on Euclidean distance. Furthermore, we carried out comprehensive experiments on three available datasets, and the results show that our proposed method is superior to the previous state-of-the-art methods with low parameter and computational requirements.

\bibliographystyle{IEEEtran}
\bibliography{refs}

\end{document}